\documentclass[graybox]{svmult}

\usepackage{type1cm}        
\usepackage{makeidx}         
\usepackage{graphicx}        
\usepackage{multicol} 
\usepackage[bottom]{footmisc}

\usepackage{booktabs}

\usepackage{newtxtext}       %
\usepackage[varvw]{newtxmath}       

\makeindex             

\begin{document}

\title*{Domain-specific Language Pre-training for Dialogue Comprehension on Clinical Inquiry-Answering Conversations}

\titlerunning{Domain-specific Pre-training for Dialogue Comprehension on Clinical Conversations}
\author{Zhengyuan Liu, Pavitra Krishnaswamy, Nancy F. Chen}
\institute{Zhengyuan Liu \at Institute for Infocomm Research (I2R), A*STAR. \email{liu\_zhengyuan@i2r.a-star.edu.sg}
\and Pavitra Krishnaswamy \at Institute for Infocomm Research (I2R), A*STAR. \email{pavitrak@i2r.a-star.edu.sg}
\and Nancy F. Chen \at Institute for Infocomm Research (I2R), A*STAR. \email{nfychen@i2r.a-star.edu.sg}}
%
%
\maketitle

\abstract*{There is growing interest in the automated extraction of relevant information from clinical dialogues. However, it is difficult to collect and construct large annotated resources for clinical dialogue tasks. Recent developments in natural language processing suggest that large-scale pre-trained language backbones could be leveraged for such machine comprehension and information extraction tasks. Yet, due to the gap between pre-training and downstream clinical domains, it remains challenging to exploit the generic backbones for domain-specific applications. Therefore, in this work, we propose a domain-specific language pre-training, to improve performance on downstream tasks like dialogue comprehension. Aside from the common token-level masking pre-training method, according to the nature of human conversations and interactive flow of multi-topic inquiry-answering dialogues, we further propose sample generation strategies with speaker and utterance manipulation. The conversational pre-training guides the language backbone to reconstruct the utterances coherently based on the remaining context, thus bridging the gap between general and specific domains. Experiments are conducted on a clinical conversation dataset for symptom checking, where nurses inquire and discuss symptom information with patients. We empirically show that the neural model with our proposed approach brings improvement in the dialogue comprehension task, and can achieve favorable results in the low resource training scenario.}

\abstract{There is growing interest in the automated extraction of relevant information from clinical dialogues. However, it is difficult to collect and construct large annotated resources for clinical dialogue tasks. Recent developments in natural language processing suggest that large-scale pre-trained language backbones could be leveraged for such machine comprehension and information extraction tasks. Yet, due to the gap between pre-training and downstream clinical domains, it remains challenging to exploit the generic backbones for domain-specific applications. Therefore, in this work, we propose a domain-specific language pre-training, to improve performance on downstream tasks like dialogue comprehension. Aside from the common token-level masking pre-training method, according to the nature of human conversations and interactive flow of multi-topic inquiry-answering dialogues, we further propose sample generation strategies with speaker and utterance manipulation. The conversational pre-training guides the language backbone to reconstruct the utterances coherently based on the remaining context, thus bridging the gap between general and specific domains. Experiments are conducted on a clinical conversation dataset for symptom checking, where nurses inquire and discuss symptom information with patients. We empirically show that the neural model with our proposed approach brings improvement in the dialogue comprehension task, and can achieve favorable results in the low resource training scenario.}

\begin{keywords}
Machine comprehension, Clinical conversation, Language pre-training
\end{keywords}

\section{Introduction}
\label{sec:introduction}
As one of the fundamental tasks in natural language processing, machine reading comprehension is fueled by avid neural modeling investigations in recent years. Given a certain textual content, the goal is to answer a series of questions based on semantic understanding. Many studies have focused on monological documents like Wikipedia \cite{rajpurkar2016squad_1} and news articles \cite{Hermann2015cnndata}, and some recent work are focusing on dialogue comprehension \cite{liu2019healthData,ma2018dialogueQA,2019Dream}. Different from passages, human-to-human dialogues are a dynamic and interactive flow of information exchange \cite{sacks1978simplest}, which are often informal, verbose, and repetitive, and this brings unique challenges for adopting document-oriented approaches on the conversational samples.

Recently, there is growing interest in automated extraction of relevant information from clinical dialogues \cite{du2019ext-sym,liu2019turn-by-turn,liu2019healthData}, in the form of dialogue comprehension. However, neural models via supervised learning usually require a certain amount of training data, and it is difficult to collect and construct large annotated resources for clinical-related tasks. Fine-tuning large-scale pre-trained language models has become a data-efficient learning paradigm, and achieves substantial improvement on various downstream tasks and applications \cite{devlin2019bert,liu2019roberta}. However, these general-purpose language backbones are trained on the online crawled data with universal objectives. Although this can provide feature-rich contextualized representations, it also limits their capability in specific domains. On the other hand, while some recent studies have proposed methods for pre-training on dialogue samples \cite{gao2020dialogueFeed,wu2020tod,zhang2020dialogpt,zhong2022-diaLM,zou2021-abs-pre}, they are focusing more on the generation of daily conversations (e.g., open-domain social chat). When adopting such generic language backbones on more targeted scenarios such as clinical-related tasks, their performance becomes sub-optimal due to the significant domain difference \cite{krishna2021soap}.

To address such challenges, in this work, we propose a domain-specific language pre-training, and adopt it to improve spoken dialogue comprehension performance on the clinical inquiry-answering conversations.
While the common methods for constructing pre-training samples (e.g., the token masking used in BERT \cite{devlin2019bert}, the infilling and deletion used in BART \cite{lewis2020bart}) proved effective for contextualized modeling, text manipulation designed for specific resources/tasks can bring further improvement \cite{gururangan2020dont-stop}. Therefore, considering the nature of human conversations \cite{sacks1978simplest} and the interactive flow of inquiry-answering dialogues, we propose a set of sample manipulations for conversational pre-training. At the token level, we introduce a masking strategy especially on speaker tokens, and we propose a permuting operation for better speaker role modeling. At the utterance level, we propose the utterance masking and intra-topic permutation scheme. More specifically, given a dialogue, we randomly select one utterance, and mask it or exchange it with another span in the same topic, and the language model is guided to reconstruct the coherent conversation flow based on the context. Moreover, we add an additional token at the beginning of each utterance to explicitly present the utterance boundary information.

Our experiments are conducted on a multi-topic symptom checking conversation dataset, where nurses inquire and discuss symptom information with patients. We empirically show that the proposed approach brings significant improvement on the dialogue comprehension task, especially in the low resource scenario.

\section{Domain-specific Language Pre-training}
\label{sec:methodology}
In this section, we elaborate on the six types of pre-training sample generation. We then describe the reconstruction-based learning process of language modeling to fuse domain-specific conversational features.

\subsection{Conversation-based Sample Construction}
Human-to-human spoken conversations are an interactive process of information exchange. Compared with the monological passages, a language backbone for multi-party dialogues is required to infuse the conversational linguistic features \cite{gu2020speaker}, such as speaker roles and utterance boundary information \cite{liu2019turn-by-turn}, as well as the underlying dialogue discourse structures \cite{sacks1978simplest}.

To guide language backbones to model the characteristics of conversations, we adopt the following six types of pre-training sample, as shown in Table \ref{table:pretrain-types}.\\
\textbf{Token Masking} Following the common pre-training scheme proposed in \cite{devlin2019bert}, for tokens in the sequence, 5\% of them are randomly sampled and replaced with a $<$mask$>$ token.\\
\textbf{Token Infilling} Following the denoising learning scheme proposed in \cite{lewis2020bart}, 5\% of input tokens are randomly sampled and replaced with random tokens extracted from the vocabulary.\\
\textbf{Speaker Masking} To encourage the model to grasp the speaker information \cite{gu2020speaker} in the interactive flow. We randomly sampled 10\% of the utterances, and replaced their speaker tokens with a $<$mask$>$ token.\\
\textbf{Speaker Permutation} Previous work shows that the text infilling scheme is helpful to tackle the training-testing inconsistency issue of masking-based methods \cite{lewis2020bart}. Thus, aside from the speaker masking, we randomly sampled 10\% of the utterances, and exchange its speaker with another one in the conversation.\\
\textbf{Utterance Masking} To encourage the model to grasp more contextual information at the utterance level, we further adopt a span masking scheme. More specifically, we randomly sampled 10\% of the utterances and mask the whole span, and the model will try to recover it based on the context understanding.\\
\textbf{Intra-Topic Utterance Permutation} While human conversations are often less structured than well-organized documents, they are inherently organized around the dialogue topics in a coarse-grained structure \cite{sacks1978simplest}. Therefore, we propose an intra-topic utterance permutation strategy, in which we randomly sampled 5\% of the utterances, and exchange them with another one in the same topic. This operation injects more noise into the conversation flow, and the model can only restore the original order by leveraging the underlying dialogue discourse information.\\
Moreover, we add a special $<$u$>$ token at the start position of each utterance, which can convey the utterance-level boundary information \cite{zhang2020dialogpt}, and is similar to the sentence-level $<$s$>$ token used in document-based approaches \cite{liu2019roberta}.

\begin{table*}[t!]
\caption{Pre-training data processing for dialogue samples. Here the raw utterances are extracted from the synthetic clinical inquiry-answering conversations. $<$mask$>$ denotes places replaced with the mask token. $<$random$>$ denotes replacing with a random token from a vocabulary. The permutation is to change the order of two randomly selected tokens/utterances.}
\begin{center}
\resizebox{1.0\linewidth}{!}{
\begin{tabular}{p{6.5cm}p{6.5cm}}
\toprule
     \textbf{Raw Utterances} & \textbf{Processed Utterances} \\
     \midrule
     \textbf{Token Masking} & \\
     Nurse: Do you have any headache at night? & Nurse: Do you have any headache at \textbf{$<$mask$>$}? \\
     Patient: No no headache, just a bit cough... & Patient: No no headache, just a \textbf{$<$mask$>$} cough.. \\
     Nurse: Cough? you mean cough at every night? & Nurse: Cough? you \textbf{$<$mask$>$} cough at every night? \\
     
    \midrule
    \textbf{Token Infilling} & \\
    Nurse: Do you have any headache at night? & Nurse: Do you have any \textbf{$<$random$>$} at night? \\
    Patient: No no headache, just a bit cough. & Patient: No \textbf{$<$random$>$} headache, just a bit cough.. \\
    Nurse: Cough? you mean cough at every night? & Nurse: Cough? you mean cough at every night? \\
    
    \midrule
     \textbf{Speaker Masking} & \\ 
     Nurse: Do you have any headache at night? & Nurse: Do you have any headache at night? \\
     Patient: No no headache, just a bit cough.. & \textbf{$<$mask$>$}: No no headache, just a bit cough.. \\
     Nurse: Cough? you mean cough at every night? & \textbf{$<$mask$>$}: Cough? you mean cough at every night? \\

    \midrule
     \textbf{Speaker Permutation} & \\
     Nurse: Do you have any headache at night? & Do you have any headache at night?  \\
     Patient: No no headache, just a bit cough.. & \textbf{Nurse}: No no headache, just a bit cough.. \\
     Nurse: Cough? you mean cough at every night? & Nurse: Cough? you mean cough at every night? \\

    \midrule
     \textbf{Utterance Masking} & \\ 
     Nurse: Do you have any headache at night? &  Nurse: Do you have any headache at night? \\
     Patient: No no headache, just a bit cough.. & Patient: \textbf{$<$mask$>$ $<$mask$>$ $<$mask$>$ $<$mask$>$}.. \\
     Nurse: Cough? you mean cough at every night? & Nurse: Cough? you mean cough at every night? \\

    \midrule
     \textbf{Intra-Topic Utterance Permutation} & \\ 
     Nurse: Do you have any headache at night? & Nurse: Do you have any headache at night?  \\
     Patient: No no headache, just a bit cough.. & \textbf{Nurse: Cough? you mean cough at every night?}  \\
     Nurse: Cough? you mean cough at every night? & \textbf{Patient: No no headache, just a bit cough..}\\
    
\bottomrule
\end{tabular}}
\end{center}
\label{table:pretrain-types}
\end{table*}

\subsection{Experiment Setup of Pre-training}
\begin{figure}[t!]
    \begin{center}
    \includegraphics[width=0.75\textwidth]{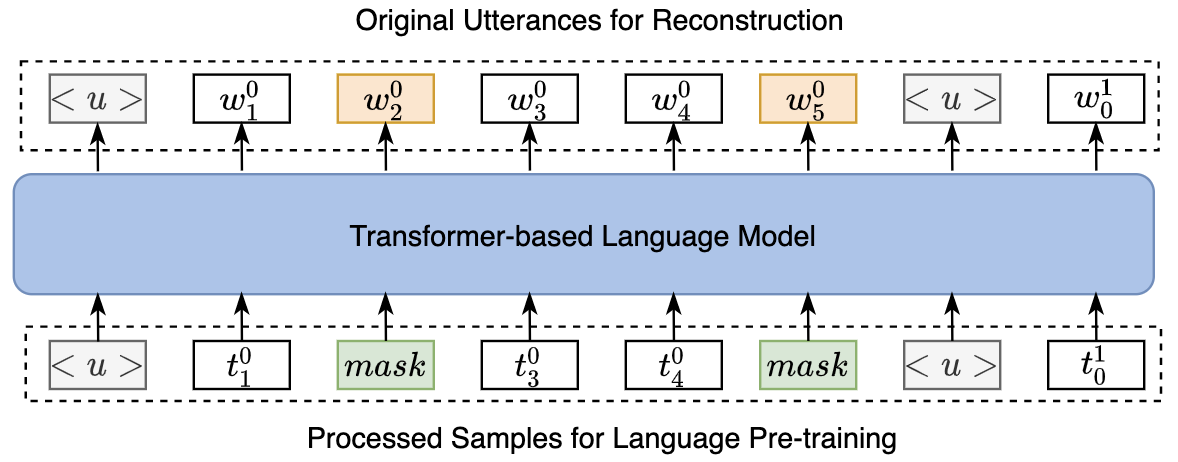}
    \end{center}
     \caption{Overview of the language model pre-training. Here we adopt the bi-directional mask-recovering training as in \cite{liu2019roberta}. Superscript values denote utterance ID.}
    \label{fig-pretrain}
\end{figure}

With the conversation samples built on the aforementioned strategies, we conduct the language backbone pre-training, and a Transformer-based neural architecture is used \cite{vaswani2017transformer}. To leverage the generic prior language knowledge, we select  \textit{`RoBERTa-base'} model to initialize the Transformer model, and conduct the \textbf{reconstruction-based} learning process \cite{liu2019roberta}. More specifically, as shown in Figure \ref{fig-pretrain}, the input sequence is the dialogue content after text manipulation, and the target sequence consists of the tokens from the original utterances.

In our experiment, the pre-training data was the combination of the SAMSum corpus \cite{gliwa2019samsum}, a subset of OpenSubtitles \cite{lison2016opensubtitles2016}, and a set of synthetic clinical symptom checking conversations (40k samples). SAMSum is a social chat corpus consisting of 16k dialogues. OpenSubtitles is compiled from a large collection of TV and movie scripts across many languages, and we randomly selected 50k samples from the English part. The mixed conversational data contains a certain amount of dialogues with multiple participants, speaker role information, and conversational structures. During training, we fixed the max input length to 512 tokens. When constructing the pre-training samples, we first conducted text manipulations on tokens and speaker entities. Then the utterance-level operations were performed. We trained the language backbone with 5,000 warm-up steps. Batch size was set to 64 via applying gradient accumulation, and the initial learning rate was set at 1e-5. Cross-entropy was used as the loss function, and we selected the checkpoints at the knee point of loss decrease.

\section{Dialogue Comprehension on Clinical Inquiry-Answering Conversations}

\subsection{Task Definition}
\label{ssec:task_definition}
One example of the dialogue comprehension task on clinical inquiry-answering conversations is shown in Table \ref{table:qa_example}. The input consists of a multi-turn symptom checking dialogue $D$ and a question $Q$ specifying a \textit{symptom} with one of its \textit{attributes}; the output is the extracted answer $A$ from the given dialogue. A training or test sample is defined as $S=\{D, Q, A\}$. Five attributes, specifying certain details of clinical significance, are defined to characterize the answer types of $A$: (1) \textit{time} the patient has been experiencing the symptom, (2) \textit{activities} that trigger the symptom (to occur or worsen), (3) \textit{extent} of seriousness, (4) \textit{frequency} occurrence of the symptom, and (5) \textit{location} of symptom. For each symptom/attribute, it can take on different linguistic expressions, defined as \textit{entities}.

\begin{table}[ht!]
\small
\caption{\label{table:qa_example} One example of the reading comprehension task on clinical inquiry-answering conversations. The synthetic dialogue is used for demonstration.}
\begin{center}
\begin{tabular}{p{9cm}}
\toprule
\textbf{Conversation Example (Truncated)} \\ 
\midrule
Nurse: Hi Mr.[Name], you were discharged on [date]. There are some questions I'd like to check with you.\\
Patient: Ok, Ok ... I think I feel better ...\\
Nurse: Is your left leg still {\color{blue}swollen}? You said so the last time I call you?\\
Patient: Yes, {\color{blue}only a bit} when I drink too much water ...\\
\midrule
\textbf{Question:} What is the extent of the swollen? \\
\midrule
\textbf{Reference Answer Span:} only a bit \\
\bottomrule
\end{tabular}
\end{center}
\end{table}

\begin{figure}[ht!]
    \begin{center}
    \includegraphics[width=0.67\textwidth]{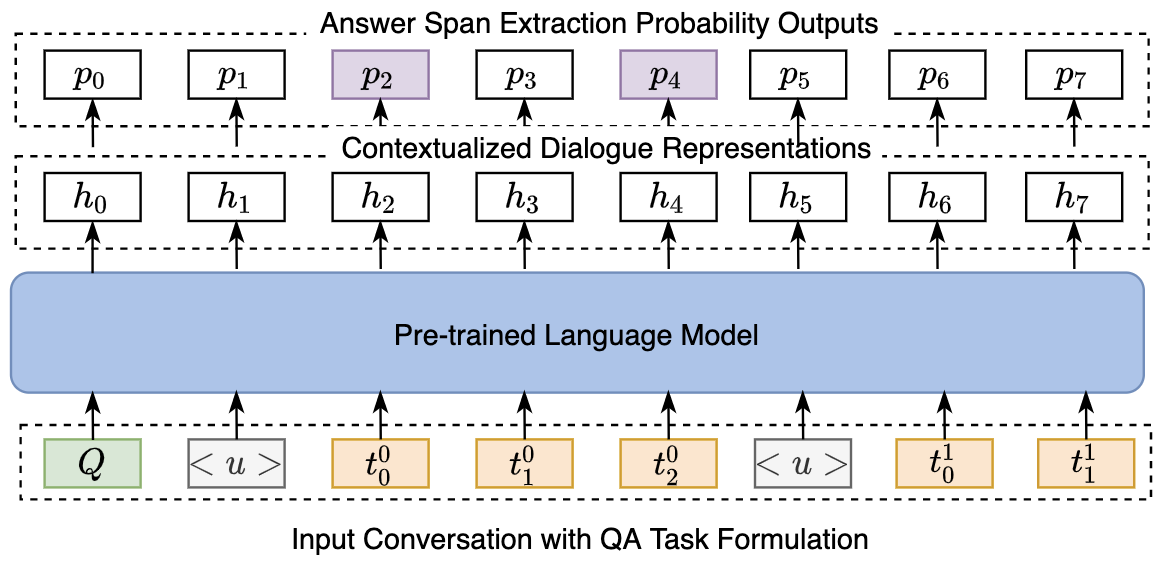}
    \end{center}
     \caption{Overview of the machine comprehension model on a question-answering task. $Q$ presents the question sequence (in green), and it is concatenated with the conversation sequence as input. Following \cite{devlin2019bert,liu2019healthData}, the answer span is extracted from the conversation content by predicting the start and end positions (in purple).}
    \label{fig-qa-model}
\end{figure}

\subsection{Clinical Dialogue Corpus}
\label{ssec:corpus}
The reading comprehension task is conducted on the data of nurse-to-patient symptom monitoring conversations.
The corpus was inspired by real dialogues in the clinical setting where nurses inquire about symptoms of patients \cite{liu2019healthData}. Linguistic structures at the semantic, syntactic, discourse, and pragmatic levels were abstracted from these conversations to construct templates for simulating multi-turn dialogues (40k samples in our settings). The informal styles of expressions, including incomplete sentences, incorrect grammar, and diffuse flow of topics were preserved.

A team of linguistically trained personnel refined, substantiated, and corrected the automatically simulated dialogues by enriching verbal expressions through different English speaking populations in Asia, Europe, and the U.S., validating logical correctness through checking if the conversations were natural, reasonable, and not disobeying common sense, and verifying the clinical content by consulting certified and registered nurses. 
These conversations cover 9 topics/symptoms (e.g. headache, cough).
For each conversation, the average word number is 255 and the average turn number is 15.5. For the comprehension task, questions were raised to query different attributes of a specified symptom; e.g., \textit{How frequently did you experience headaches?} Answer spans in the dialogues were labeled with start and end indices, following the annotation scheme as in \cite{rajpurkar2016squad_1}. Note that if the queried symptom or attribute is not mentioned in the dialogue, the ground-truth output is \textit{``No Answer''}, as the same definition in \cite{liu2019healthData}.

\subsection{Baseline Models}
We further fine-tuned the Transformer-based model on the dialogue comprehension task, and compared it with several baselines, including \textbf{Pointer LSTM} \cite{wang2016matchLSTM}, Bi-Directional Attention Flow network (\textbf{Bi-DAF}) \cite{seo2016bidaf}, and \textbf{R-Net} \cite{wang2017rnet}.
To evaluate the effectiveness of our domain-specific language pre-training, we use the \textbf{Vanilla Transformer} and the original \textbf{RoBERTa-base} model as control, and our proposed model is 
\textbf{RoBERTa-base w/ Domain-specific Pre-training}. As shown in Figure \ref{fig-qa-model}, we formulate the comprehension task as an answer extraction process. With the feature-rich contextualized representation, the answer span is generated by predicting its start/end position in the sequence, by adding a linear layer on the last layer hidden states from the language modeling \cite{devlin2019bert,liu2019roberta}.

\subsection{Training Configuration}
All models were implemented with Pytorch and Hugging Face Transformers \cite{paszke2019pytorch}.
For models without pre-trained language backbones (e.g. Bi-DAF, R-Net), Glove embedding \cite{pennington2014glove} was utilized, and out-of-vocabulary words were replaced with the \textit{$<$unk$>$} token. Hidden size and embedding dimension were 300, and those of Transformer-based models were 768. We used Adam \cite{kingma2014adam} with batch size 32, and gradient accumulation was applied. The initial learning rates were set at 2e-5, and dropout rate \cite{srivastava2014dropout} was set to 0.2. During training, the validation-based early stop strategy was applied. During prediction, we selected answer spans using the maximum product of $p_{start}$ and $p_{end}$.

\subsection{Evaluation: Comparison with Baselines}
\label{ssec:basic_result}
We conduct the evaluation on the synthetic clinical dialogue corpus, where the training, validation, and test size were 40k, 3k, and 3k, respectively. We adopted Exact Match (EM) and F1 score as metrics as the SQuAD benchmark \cite{rajpurkar2016squad_1}.

As shown in Table \ref{result-baseline}, the vanilla Transformer model obtains a slightly lower performance than the non-Transformer strong baselines (i.e. Bi-DAF and R-Net), and \textit{`RoBERTa-base'} is on par with them. This demonstrates that the prior knowledge from general language pre-training is beneficial for the downstream tasks. With the conversational pre-training, our proposed mode obtains substantial gains and achieve the best EM and F1 scores, showing that the domain-specific feature fusion is effective.

\begin{table}[t!]
\small
\caption{\label{result-baseline}
Evaluation result of the baseline models and our approach on the test set. \textit{Domain PT} denotes the proposed domain-specific pre-training.}
\begin{center}
\begin{tabular}{p{5.5cm}cc}
\toprule
\textbf{Model} & \textbf{\ \ EM Score\ \ } & \textbf{\ \ F1 Score\ \ } \\ 
\midrule
Pointer LSTM & 77.81 & 82.71 \\
Bi-Attention Flow (Bi-DAF) & 87.31 & 88.57 \\
R-Net (Our Implementation) & 88.22 & 90.13 \\
\midrule
Vanilla Transformer & 85.38 & 85.92 \\
RoBERTa-base w/o Domain PT & 88.37 & 90.15 \\
RoBERTa-base w/ Domain PT & \textbf{92.31} & \textbf{93.69} \\
\bottomrule
\end{tabular}
\end{center}
\end{table}

\subsection{Evaluation in Low-Resource Scenarios}
\label{ssec:low_res_result}
\begin{figure}[t!]
    \begin{center}
    \includegraphics[width=0.7\textwidth]{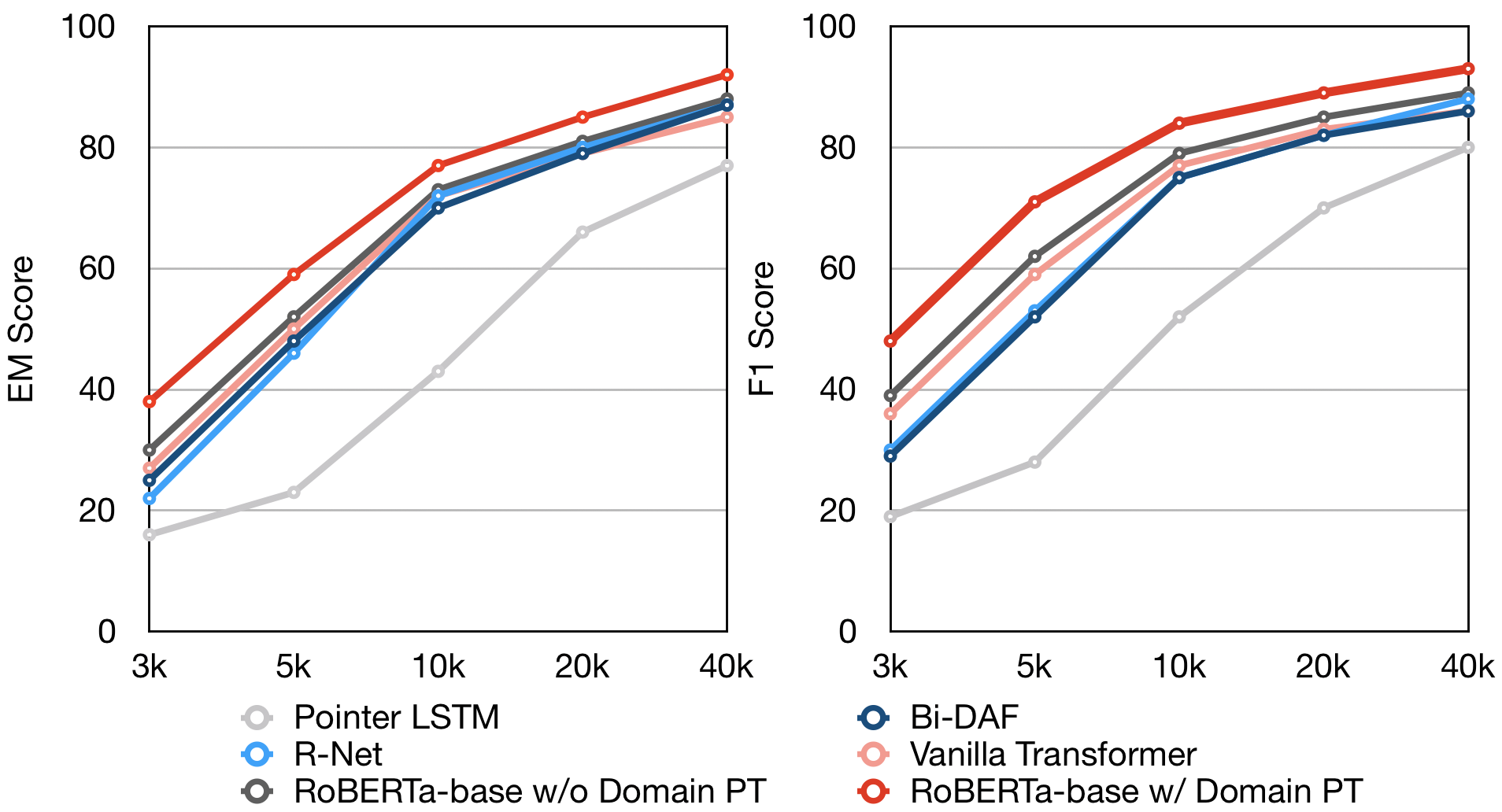}
    \end{center}
     \caption{\label{fig-low-resource} Experimental result on the low-resource training. X axis is the sample size, Y axis is the evaluation metrics, including exact match (EM) and F1 score.}
\end{figure}

The limited amount of training data is a major pain point for clinical-related language tasks, as it is time-consuming and labor-intensive to collect and annotate the corpus at a large scale. Following the observation in previous work \cite{gururangan2020dont-stop}, we expect the domain-specific language modeling can result in more efficient learning on downstream tasks. To simulate the low-resource training scenario, we conducted experiments on a range of smaller training sizes (from 3k to 40k) with a fixed-size test set (3k samples). As shown in Figure \ref{fig-low-resource}, the proposed approach outperforms all other models significantly, especially when the training size is smaller than 20k.

\subsection{Evaluation: Pre-training Scheme Comparison}
\label{ssec:pretrain_type_exp}
To evaluate the effectiveness of the aforementioned strategies of pre-training sample construction. We conduct an experiment by adding different text manipulations to train the general-purpose language backbone. As shown in Table \ref{result-pt-table}, we observed that the token-level infilling can bring certain improvements, and introducing the conversation-related manipulations (i.e., speaker and utterance masking and permutation) are helpful for the final dialogue comprehension performance.

\begin{table}[t!]
\small
\begin{center}
\begin{tabular}{p{6.2cm}cc}
\toprule
\textbf{Model} & \textbf{ EM Score } & \textbf{ F1 Score } \\ 
\midrule
Pre-training on the \textit{RoBERTa-base} backbone \\
+ Token-level Masking & 88.79 & 90.30 \\
+ Token-level Infilling & 90.73 & 92.12 \\
+ Speaker Mask \& Permutation & 91.01 & 92.41 \\
+ Utterance Mask \& Permutation & \textbf{92.31} & \textbf{93.69} \\
\bottomrule
\end{tabular}
\end{center}
\caption{\label{result-pt-table}Performance comparison on pre-training schemes. The text manipulations are added from token level to utterance level.}
\end{table}

\section{Conclusions}
In this paper, we introduced a domain-specific language pre-training approach, and adopted it to improve   performance on downstream tasks such as question answering. Based on the linguistic characteristics of spoken dialogues, we proposed a combination of six strategies to build samples for conversational language pre-training, and conducted reading comprehension experiments on a multi-topic inquiry-answering conversation data. The experimental results showed that the proposed approach can boost performance and achieve more efficient learning outcomes. Future work include extending conversational pre-training to other clinical tasks \cite{Kurisinkel2021report,liu2019conv_summ} and resources.

\begin{acknowledgement}
Research efforts were supported by funding and infrastructure from A*STAR, Singapore (Grant No. IAF H19/01/a0/023). We gratefully acknowledge valuable inputs from Angela Ng, Hong Choon Oh, Sharon Ong, Sheldon Lee, Weiliang Huang, and Ying Zi Oh at the Department of Cardiology, Health Management Unit, and Department of Health Services Research, Changi General Hospital, Singapore. We thank the anonymous reviewers for their precious feedback to help improve and extend this piece of work.
\end{acknowledgement}

%

%
%
%
\biblstarthook{}

\end{document}